%% file: aaai24.tex
\documentclass[letterpaper]{article} 
\usepackage{aaai24}  
\usepackage{times}  
\usepackage{helvet}  
\usepackage{courier}  
\usepackage[hyphens]{url}  
\usepackage{graphicx} 
\usepackage{natbib}  
\usepackage{caption} 
\usepackage{algorithm}
\usepackage{algorithmic}

\usepackage{xcolor}
\usepackage{pifont}
\usepackage{enumitem}

\usepackage{microtype}
\usepackage[utf8]{inputenc}
\usepackage{inconsolata}

\usepackage{booktabs}
\usepackage{longtable}
\usepackage{lscape}
\usepackage{comment}
\usepackage{tabularx}
\usepackage{multirow}
\usepackage{float}
\usepackage{enumitem}

\usepackage{newfloat}
\usepackage{listings}
\DeclareCaptionStyle{ruled}{labelfont=normalfont,labelsep=colon,strut=off} 
\floatstyle{ruled}
\newfloat{listing}{tb}{lst}{}
\floatname{listing}{Listing}
\title{RESTORE: Graph Embedding Assessment Through Reconstruction}
\author{
    Hong Yung Yip\textsuperscript{\rm 1}, Chidaksh Ravuru\textsuperscript{\rm 2}, Neelabha Banerjee\textsuperscript{\rm 3}, Shashwat Jha\textsuperscript{\rm 4}, \\ Amit Sheth\textsuperscript{\rm 1}, Aman Chadha\textsuperscript{\rm 5, 6}\footnote{Work does not relate to position at Amazon.}, Amitava Das\textsuperscript{\rm 1}
}
\affiliations{
    \textsuperscript{\rm 1}Artificial Intelligence Institute, University of South Carolina \\
    \textsuperscript{\rm 2}Indian Institution of Technology, Dharwad \\
    \textsuperscript{\rm 3}Christ University, Bengaluru \\
    \textsuperscript{\rm 4}Birla Institute of Technology, Mesra \\
    \textsuperscript{\rm 5}Stanford University, California \\
    \textsuperscript{\rm 6}Amazon Alexa AI, California \\

}

\usepackage{bibentry}

\begin{document}

\maketitle


\begin{abstract}
Following the success of Word2Vec embeddings, graph embeddings (GEs) have gained substantial traction. GEs are commonly generated and evaluated extrinsically on downstream applications, but intrinsic evaluations of the original graph properties in terms of topological structure and semantic information have been lacking. Understanding these will help identify the deficiency of the various families of GE methods when vectorizing graphs in terms of preserving the relevant knowledge or learning incorrect knowledge. To address this, we propose RESTORE, a framework for intrinsic GEs assessment through graph reconstruction. We show that reconstructing the original graph from the underlying GEs yields insights into the relative amount of information preserved in a given vector form. We first introduce the graph reconstruction task. We generate GEs from three GE families based on factorization methods, random walks, and deep learning (with representative algorithms from each family) on the CommonSense Knowledge Graph (CSKG). We analyze their effectiveness in preserving the (a) topological structure of node-level graph reconstruction with an increasing number of hops and (b) semantic information on various word semantic and analogy tests. Our evaluations show deep learning-based GE algorithm (SDNE) is overall better at preserving (a) with a mean average precision (mAP) of 0.54 and 0.35 for 2 and 3-hop reconstruction respectively, while the factorization-based algorithm (HOPE) is better at encapsulating (b) with an average Euclidean distance of 0.14, 0.17, and 0.11 for 1, 2, and 3-hop reconstruction respectively. The modest performance of these GEs leaves room for further research avenues on better graph representation learning.
\end{abstract}

\input{1_introduction}
\input{2_reconstruction}

\input{3_results}

\input{4_evaluation}
\input{5_results_dis}
\input{6_discussion_futurework}
\input{7_conclusion}

\bibliography{aaai24}

\section{Appendix}
\begin{figure*}[t]
\centering
\fbox{\includegraphics[width=0.7\textwidth]{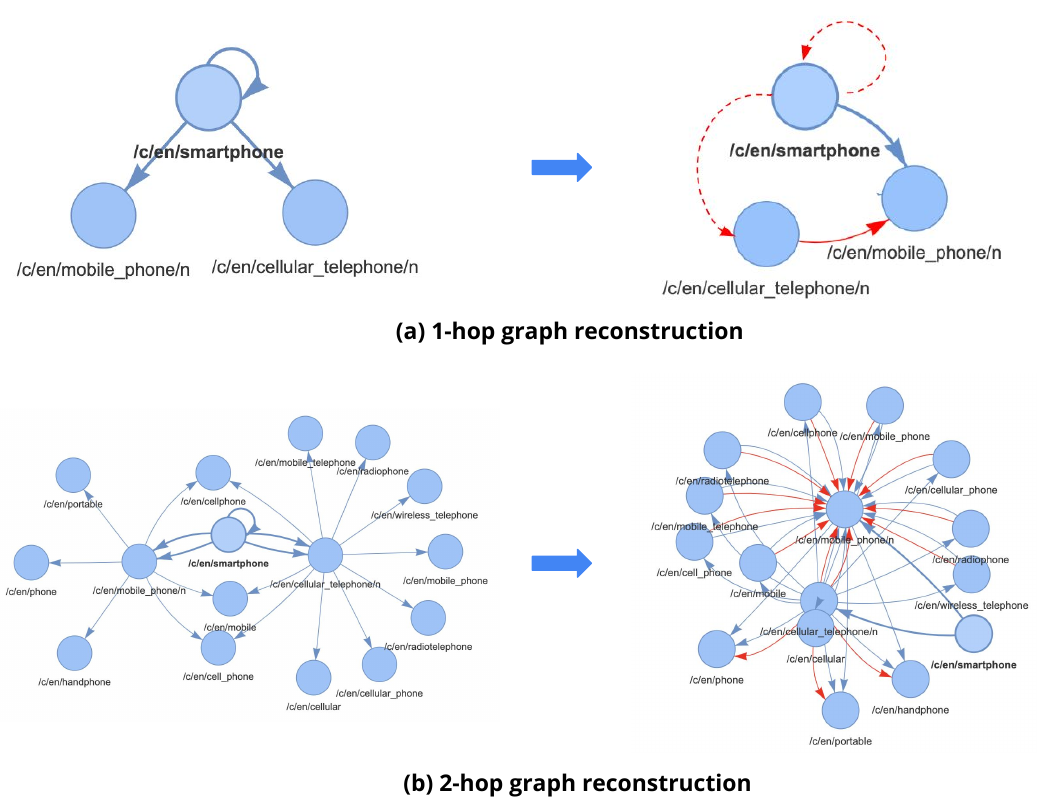}} 

\caption{\textbf{1 and 2-hop graph reconstructions with GEs from Node2Vec.} The graph on the left represents the original graph and the graph on the right represents the reconstructed graph. Red dotted lines indicate missing edges and red solid lines indicate added edges compared to the original graph.}
\label{recon_3}
\end{figure*}

\begin{figure*}[t]
\centering
\fbox{\includegraphics[width=0.7\textwidth]{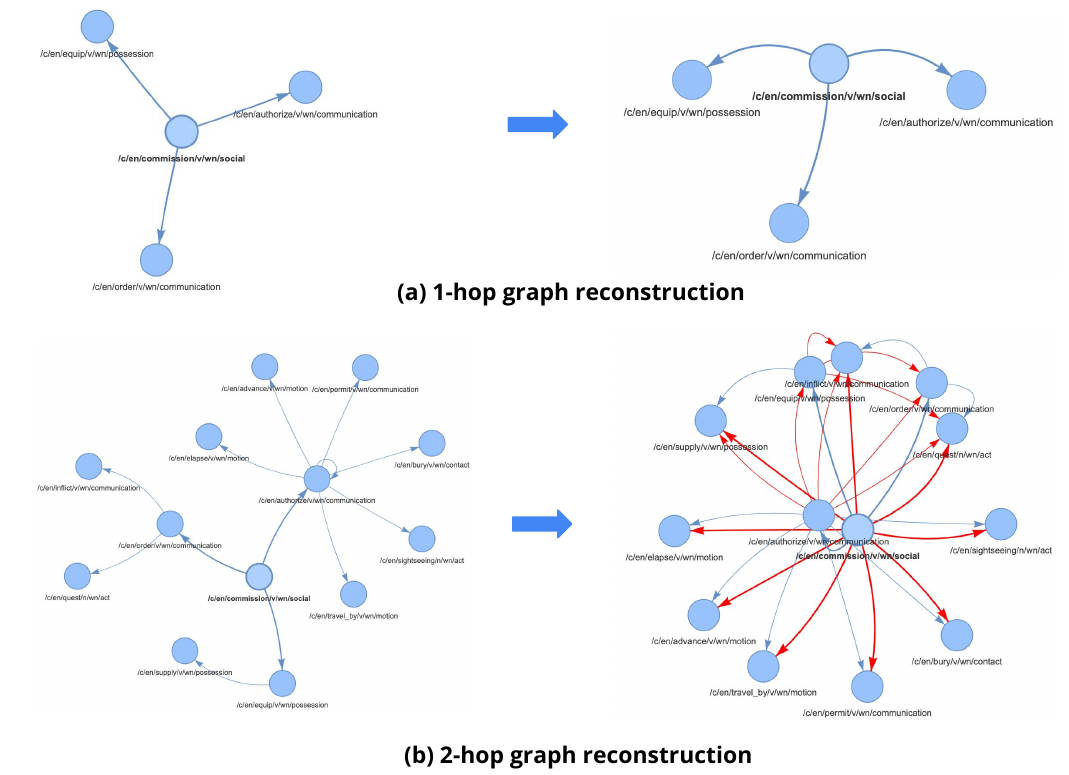}} 

\caption{\textbf{1 and 2-hop graph reconstructions with GEs from HOPE.} The graph on the left represents the original graph and the graph on the right represents the reconstructed graph. Red dotted lines indicate missing edges and red solid lines indicate added edges compared to the original graph.}
\label{recon_4}
\end{figure*}

\begin{figure*}[t]
\centering
\fbox{\includegraphics[width=0.7\textwidth]{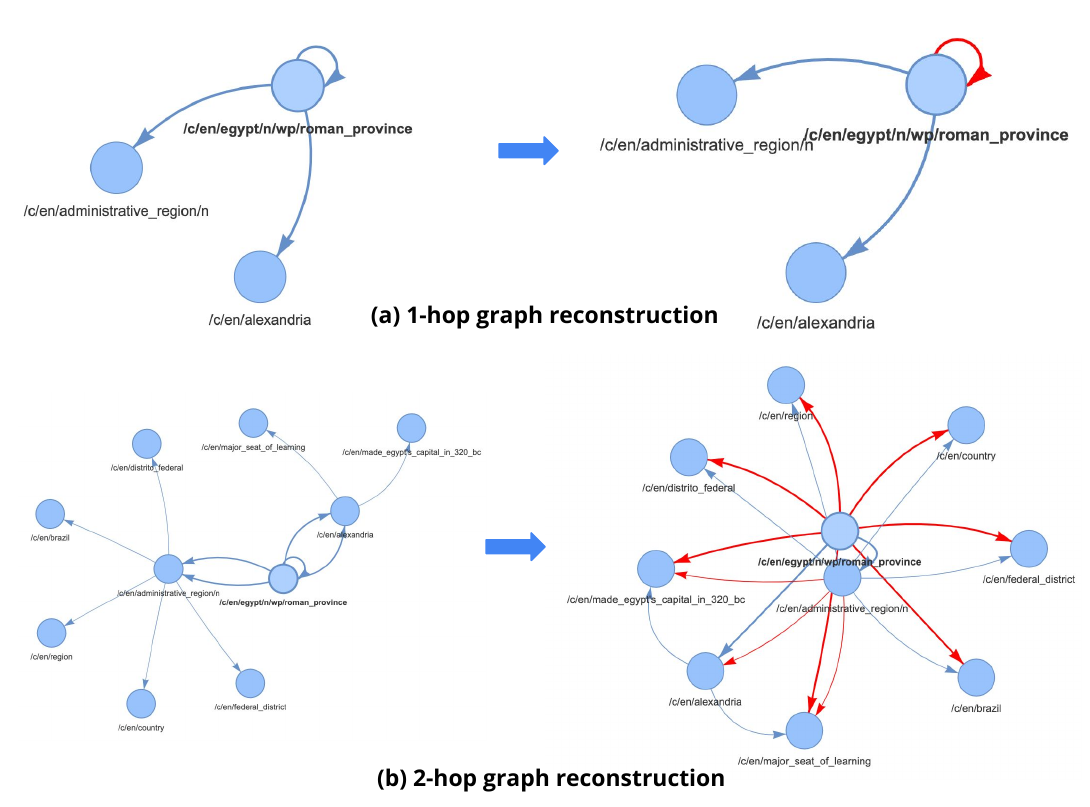}} 

\caption{\textbf{1 and 2-hop graph reconstructions with GEs from SDNE.} The graph on the left represents the original graph and the graph on the right represents the reconstructed graph. Red dotted lines indicate missing edges and red solid lines indicate added edges compared to the original graph.}
\label{recon_5}
\end{figure*}

\begin{figure*}[t]
\centering
\fbox{\includegraphics[width=0.7\textwidth]{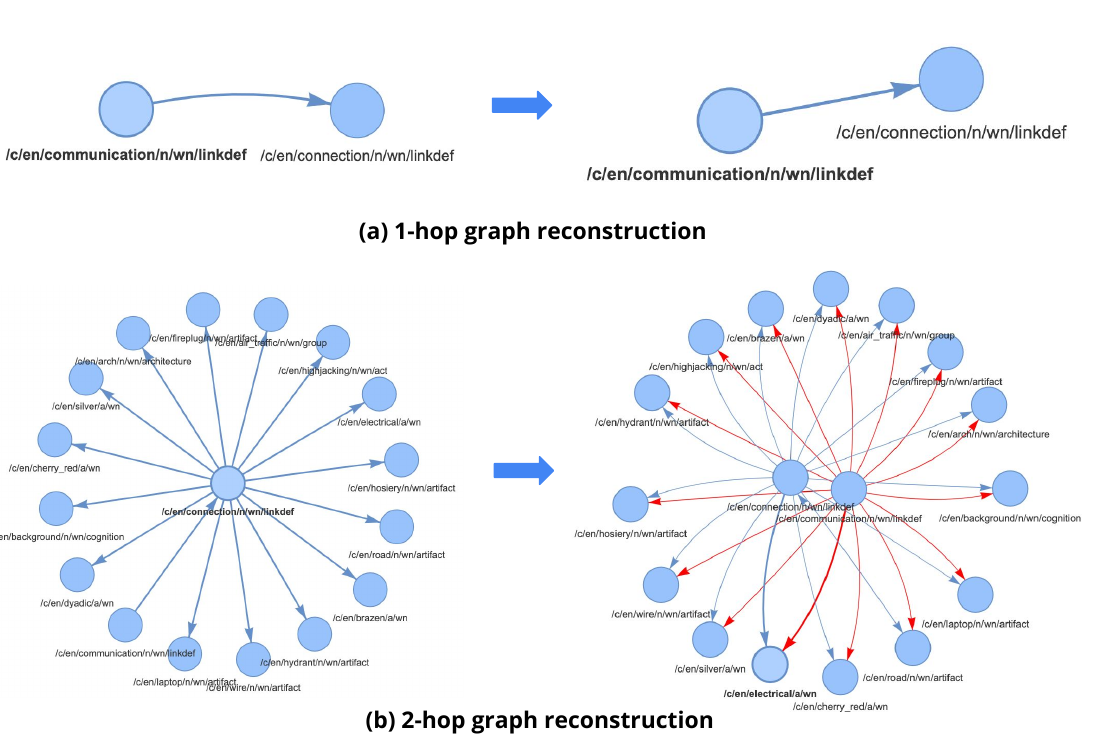}} 

\caption{\textbf{1 and 2-hop graph reconstructions with GEs from LLE.} The graph on the left represents the original graph and the graph on the right represents the reconstructed graph. Red dotted lines indicate missing edges and red solid lines indicate added edges compared to the original graph.}
\label{recon_6}
\end{figure*}

\begin{figure*}[t]
\centering
\fbox{\includegraphics[width=0.7\textwidth]{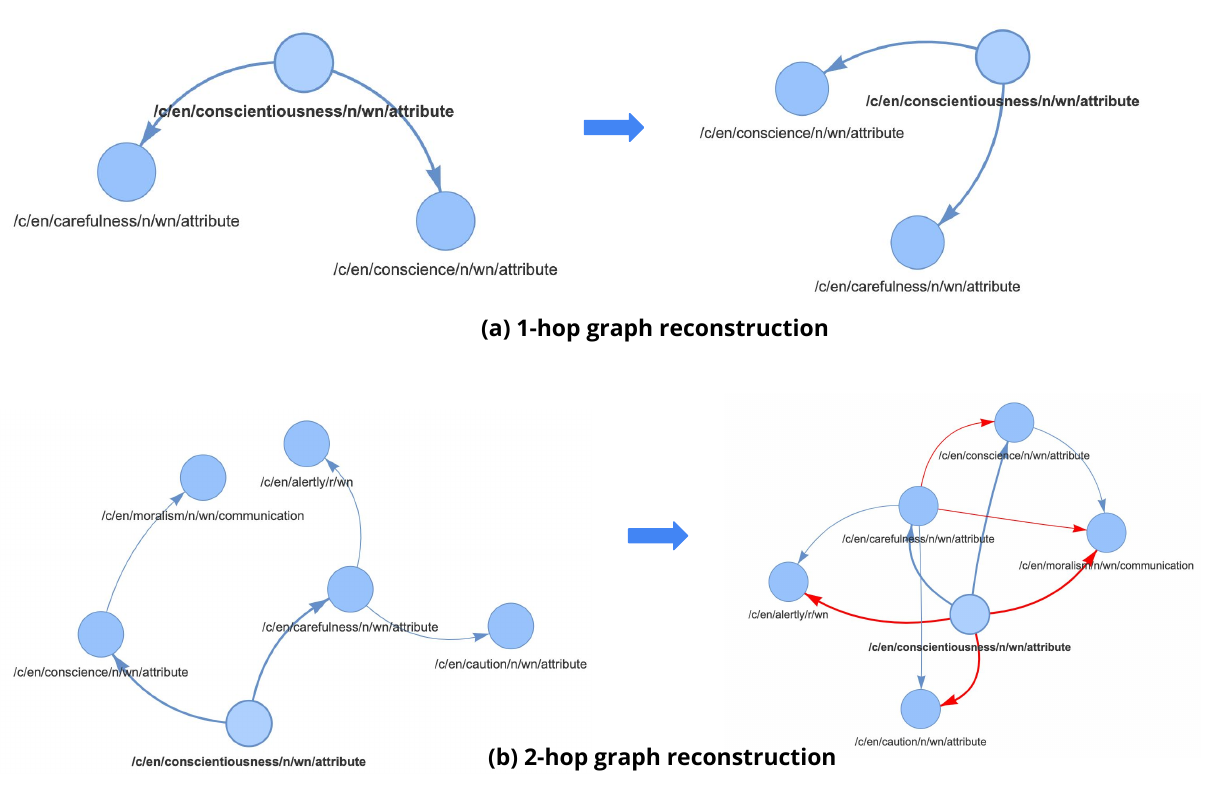}} 

\caption{\textbf{1 and 2-hop graph reconstructions with GEs from LAP.} The graph on the left represents the original graph and the graph on the right represents the reconstructed graph. Red dotted lines indicate missing edges and red solid lines indicate added edges compared to the original graph.}
\label{arch}
\end{figure*}

\begin{figure*}[t]
\centering
\fbox{\includegraphics[width=0.7\textwidth]{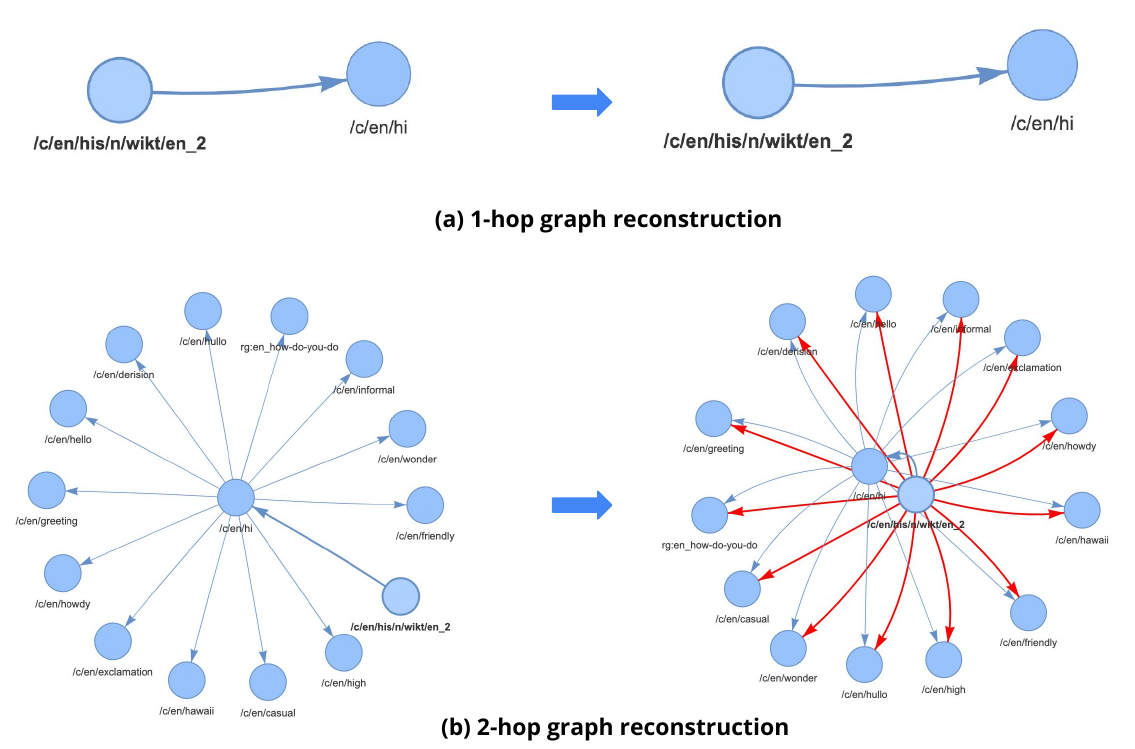}} 

\caption{\textbf{1 and 2-hop graph reconstructions with GEs from Node2Vec.} The graph on the left represents the original graph and the graph on the right represents the reconstructed graph. Red dotted lines indicate missing edges and red solid lines indicate added edges compared to the original graph.}
\label{recon_7}
\end{figure*}

\begin{figure*}[t]
\centering
\fbox{\includegraphics[width=0.8\textwidth]{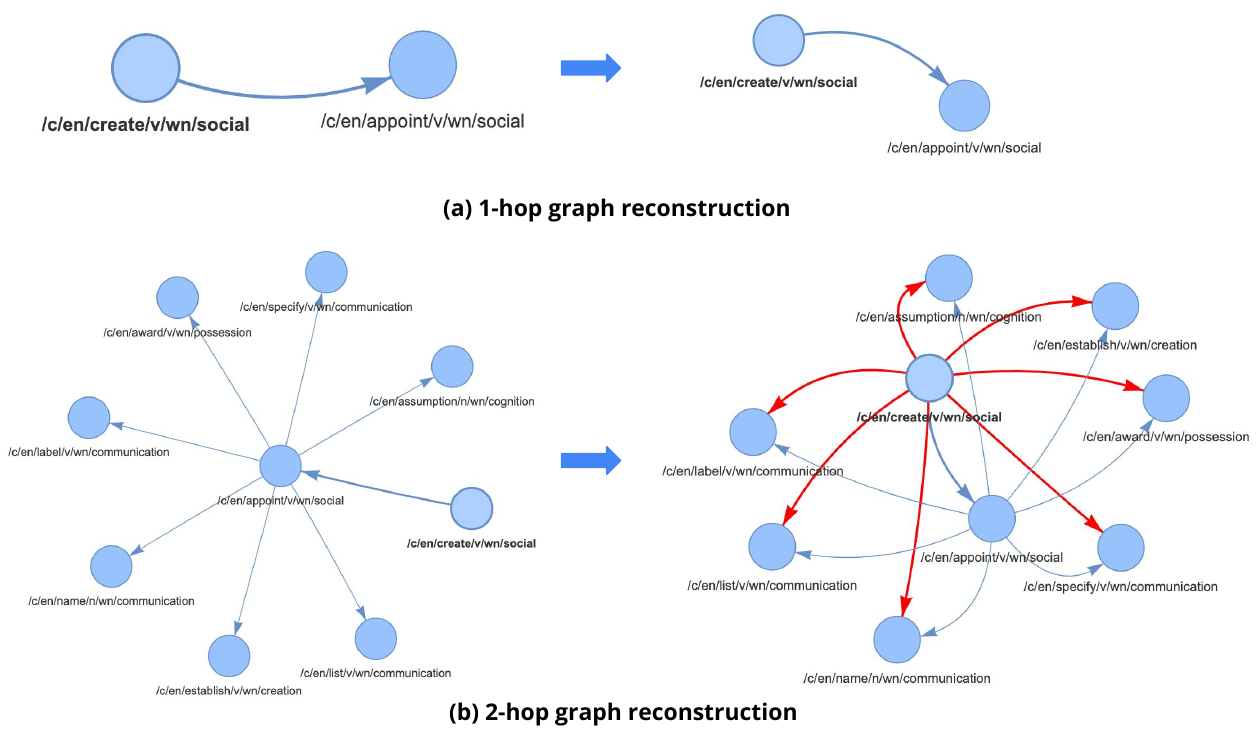}} 

\caption{\textbf{1 and 2-hop graph reconstructions with GEs from HOPE.} The graph on the left represents the original graph and the graph on the right represents the reconstructed graph. Red dotted lines indicate missing edges and red solid lines indicate added edges compared to the original graph.}
\label{recon_8}
\end{figure*}

\begin{figure*}[t]
\centering
\fbox{\includegraphics[width=0.8\textwidth]{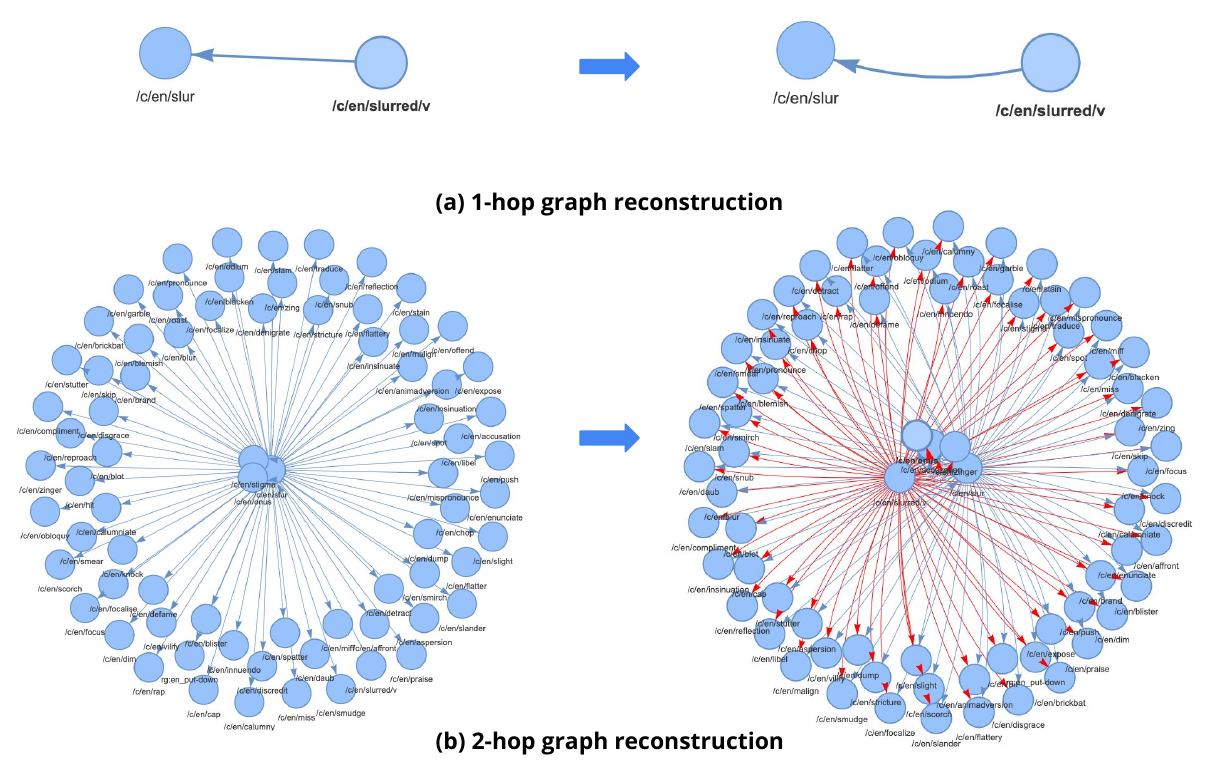}} 

\caption{\textbf{1 and 2-hop graph reconstructions with GEs from SDNE.} The graph on the left represents the original graph and the graph on the right represents the reconstructed graph. Red dotted lines indicate missing edges and red solid lines indicate added edges compared to the original graph.}
\label{recon_9}
\end{figure*}

\begin{figure*}[]
\centering
\fbox{\includegraphics[width=0.8\textwidth]{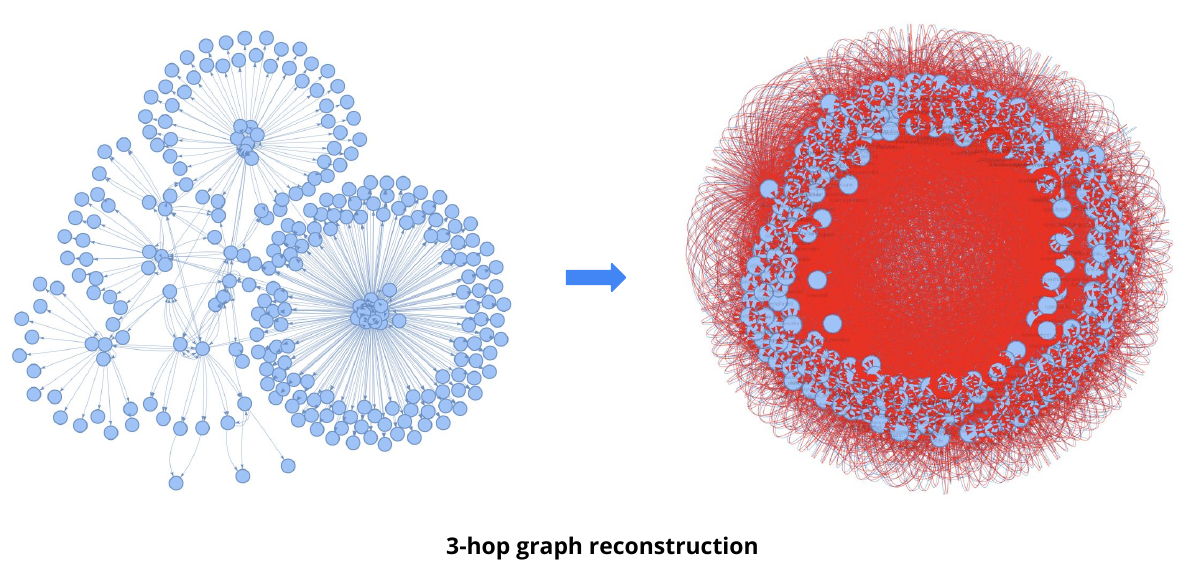}} 

\caption{\textbf{3-hop graph reconstruction with GEs from Node2Vec.} The graph on the left represents the original graph and the graph on the right represents the reconstructed graph. Red dotted lines indicate missing edges and red solid lines indicate added edges compared to the original graph.}
\label{recon_2}
\end{figure*}

Here, we attach additional examples of 1 and 2-hop graph reconstructions (Figure \ref{recon_3}-\ref{recon_9}) from various families of GE algorithms. Due to the size of the 3-hop graph, we show only 1 example of the 3-hop reconstruction as all other examples follow similar illustrations in Figure \ref{recon_2}.

\end{document}

%% file: 1_introduction.tex
\section{How effective is graph embedding in preserving both graph topology and semantic information when transforming a graph into a vector?}
An embedding is a mapping of a discrete set of objects to a continuous vector space. Embeddings have been successful in providing effective features for many neural network models. A major attraction of vector space representation is that they represent the objects in a way that captures the relationships and patterns within the data from large unannotated corpora. For example, in Natural Language Processing (NLP), a word embedding maps words in a vocabulary to vectors in a continuous vector space such that semantically similar words are close together. As the use of graph representation rises, graph embedding has gained traction. In graph analysis, a graph embedding (GE) maps nodes in a graph to vectors in a continuous space such that nodes that are structurally similar or connected in the graph are close together in the vector space.

Vector representations, however, are linguistically opaque and non-interpretable for a human. While NLP word analogies such as the \textit{king-queen} analogy popularized by Word2Vec \cite{mikolov2013efficient}, lend themselves as a standard practice for the intrinsic evaluation of word embeddings, there is a lack of consensus or standard for the intrinsic evaluation of GEs. While there are a plethora of GE algorithms have been proposed \cite{ahmed2013distributed, tang2015line, wang2016structural, ou2016asymmetric, belkin2001laplacian, perozzi2014deepwalk, grover2016node2vec, kipf2016semi, roweis2000nonlinear} and evaluated based on various graph analytic tasks, obtaining a vector representation of each node of a graph that preserves the global structure of the graph and the local connections between individual nodes is challenging. Suppose we assume the performance improvement based on extrinsic evaluations by current deep neural networks on downstream applications is attributed to the use of GEs \cite{makarov2021survey}, in that case, it is imperative that the GEs are preserving the right graph structure and semantics. As discussed in \cite{xucloser, liu2019much}, the embeddings generated by the current state-of-the-art approaches can only preserve part of the topological structure. While \cite{bollegala2018learning} have shown that by combining different source \textit{(word)} embeddings into a coherent common meta-embedding space, the generated meta-embedding is able to produce a more accurate and complete representation, the GEs counterpart is yet to be studied.

Orthogonally, generating GEs from large graphs is computationally expensive due to their size and real-world complexity \cite{fu2021ts, goyal2018graph}. Extracting only subgraphs relevant to the entities or communities of interest (domain-specific) is generally the strategy for learning semantically relevant embeddings \cite{fu2021ts}. However, determining the amount of knowledge to extract (i.e., number of hops) for a given task is non-trivial \cite{ribeiro2021survey}. The degree to which the existing GE algorithms preserve the graph properties based on the number of hops is yet to be investigated.

\begin{figure*}[t]
\centering
\includegraphics[width=\textwidth]{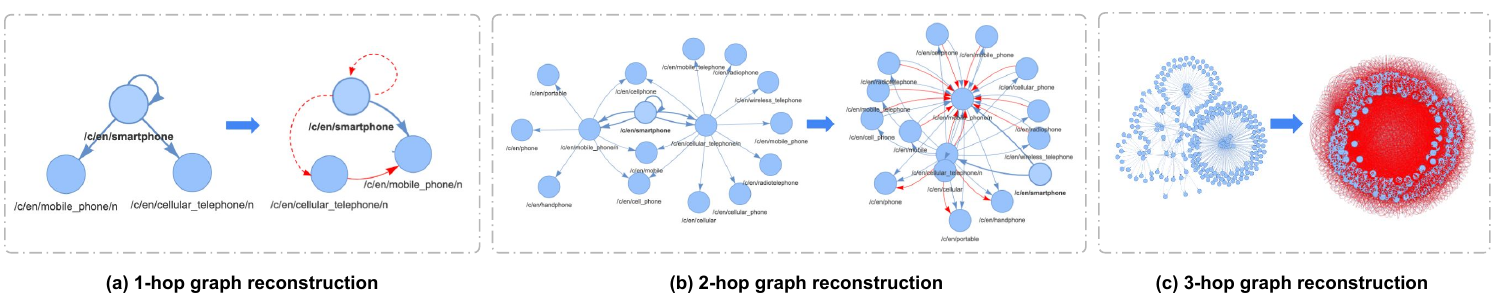} 
\caption{\textbf{CSKG Subgraphs Reconstruction.} In this illustration example, we generate the 1, 2, and 3-hop subgraphs for the node of interest (/c/en/smartphone; c: ConceptNet, en: English) and train the corresponding GEs with \textit{Node2Vec}. In each bounding box, the graph on the left represents the original graph and the graph on the right represents the reconstructed graph. Red dotted lines indicate missing edges and red solid lines indicate added edges compared to the original graph.}
\label{recon}
\end{figure*}

\section{Our Contributions}
\label{sec:Contri}
The proposed RESTORE is an evaluation framework to assess the quality of embeddings generated for a given node by different families of GE algorithms through graph reconstruction along three dimensions:

\begin{enumerate}
  \item The \textbf{type} of graph properties preserved: (a) topological structure and (b) semantic information between nodes.
  \item Which \textbf{family} of GE algorithm(s) are \textbf{better} at preserving 1(a) versus 1(b).
  \item The degree of information that is preserved (retained, added, and missed) by the various GE algorithms with increasing \textbf{number of hops}.
\end{enumerate}

It is difficult to determine which GE is capable and appropriate for a specific downstream application based on the aforementioned dimensions without exhaustive experimentation. Through RESTORE, we intend to shed light on the performance of each GE family through comparative analysis. We omit the typed relations between nodes and thereof their reconstruction assessment as the task falls under the extreme classification category and beyond the scope of this paper. 
Our contributions can be summarized as follows:

\begin{itemize}
\item Understand the effectiveness of graph embeddings in retaining graph topological structure and/or semantic information with increasing graph size (number of hops).
\item Reconstruct the original graph from the vector produced by graph embedding methods to understand the degree of topological/semantic information captured.
\item Assess various graph embedding algorithms using word semantic and analogy datasets (including vector arithmetic such as the \textit{king - man + woman = queen} analogy, a standard in NLP).
\end{itemize}

The paper is organized as follows. We first provide the preliminaries required to understand GE and introduce the three families of GE algorithms. We then describe the graph reconstruction task and our experimental setup in detail. Next, we assess the topological structure reconstruction accuracy and evaluate the degree of semantic information retained by the different GE algorithms with various word semantic and analogy tests. Finally, we discuss the findings and limitations of this work and draw our conclusions.

%% file: 2_reconstruction.tex
\begin{figure*}[t]
\centering
\includegraphics[width=0.9\textwidth]{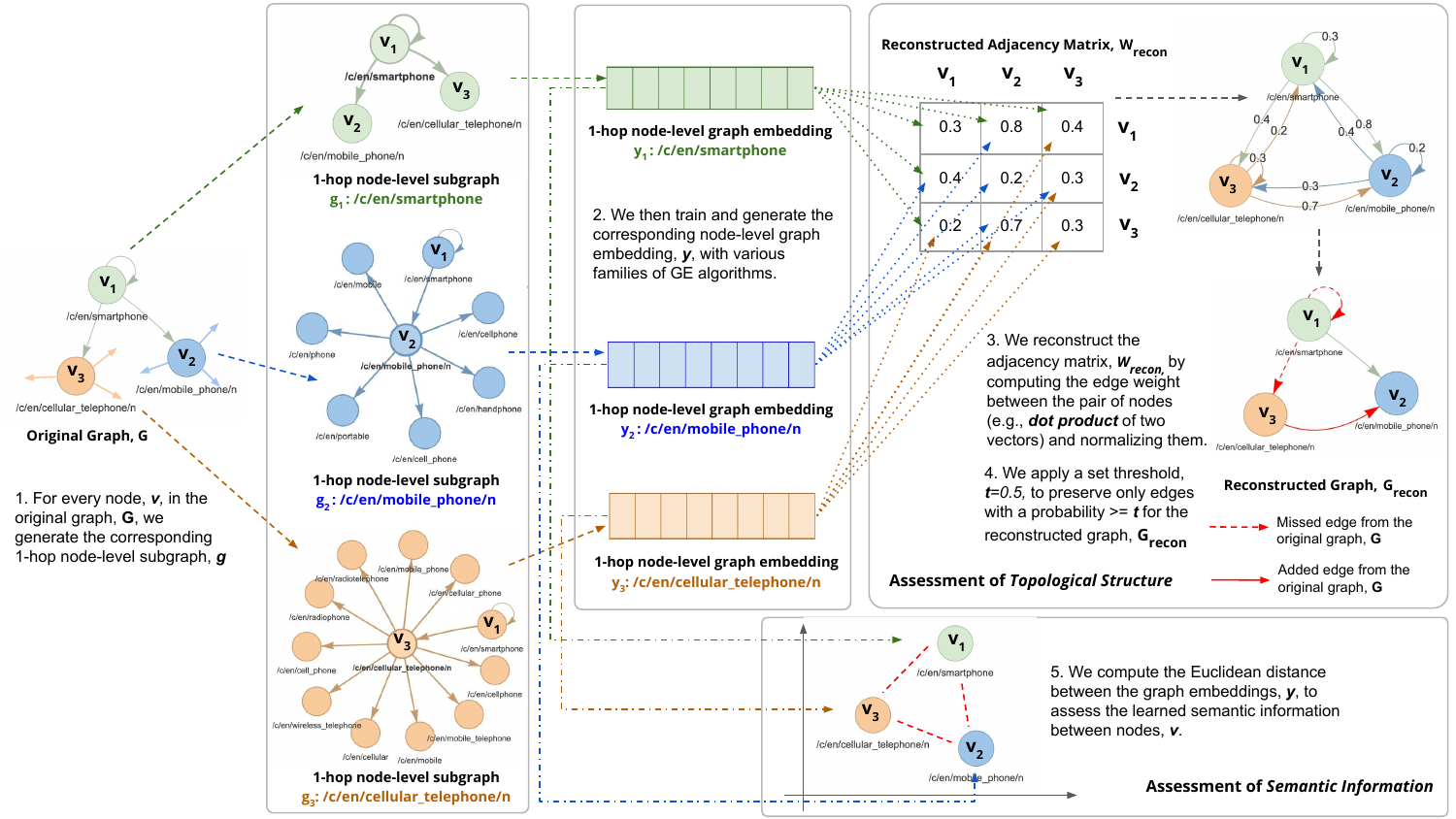} 

\caption{\textbf{The RESTORE framework to assess the degree of topological structure and semantic information preserved by GEs during graph reconstruction.} In this illustration example, our node of interest is ($v_1$: /c/en/smartphone). We first generate the corresponding 1-hop node-level subgraphs, \textbf{\textit{$g_i$}}, by accounting all in and out-degrees, for every immediate 1-hop neighbors of $v_1$ ($v_2$:  /c/en/mobile\_phone/n; $v_3$: /c/en/cellular\_telephone/n; c: ConceptNet; en: English). We then generate the corresponding subgraph embedding, \textbf{\textit{$y_i$}}, with various families of GE algorithms. Thereafter, we reconstruct the adjacency matrix, \textbf{\textit{$w_{recon}$}} by computing the pairwise edge weight between nodes (e.g., dot product of two vectors) and normalizing them. A set threshold, \textit{t} is applied to preserve only edges with a probability $\geq$ \textit{t} for the reconstructed graph. To assess the learned semantic information between nodes, \textit{V}, we compute the pairwise Euclidean distance between the GEs. We iterate the same process for 2-hop and 3-hop graph reconstructions for the node of interest.}
\label{arch}
\end{figure*}

\begin{table*}[h]
\begin{center}
\scalebox{0.75}{
\begin{tabular}{lcccccc}
\toprule
\multicolumn{7}{c}{\textbf{Commonsense Knowledge Graph (CSKG)} \cite{ilievski2021cskg}}                                                                                                                        \\ 
\toprule
\multicolumn{1}{c}{\textit{V}}                 & \multicolumn{6}{c}{2,160,968}                                                                                                                                                                                                 \\ 
\multicolumn{1}{c}{\textit{E}}                 & \multicolumn{6}{c}{6,001,531}                                                                                                                                                                                                 \\ 
\toprule
\multicolumn{1}{l}{\textbf{Properties}} & \multicolumn{1}{c}{\textbf{Min. \textit{V}}} & \multicolumn{1}{c}{\textbf{Avg. \textit{V}}} & \multicolumn{1}{c}{\textbf{Max. \textit{V}}} & \multicolumn{1}{c}{\textbf{Min. \textit{E}}} & \multicolumn{1}{c}{\textbf{Avg. \textit{E}}} & \multicolumn{1}{c}{\textbf{Max. \textit{E}}} \\ \toprule
\multicolumn{1}{l}{1-hop}               & \multicolumn{1}{c}{1}                 & \multicolumn{1}{c}{4}                 & \multicolumn{1}{c}{111}               & \multicolumn{1}{c}{1}                 & \multicolumn{1}{c}{3}                 & 110               \\ 
\multicolumn{1}{l}{2-hop}               & \multicolumn{1}{c}{1}                 & \multicolumn{1}{c}{215}               & \multicolumn{1}{c}{10,287}            & \multicolumn{1}{c}{1}                 & \multicolumn{1}{c}{250}               & 21,395            \\ 
\multicolumn{1}{l}{3-hop}               & \multicolumn{1}{c}{1}                 & \multicolumn{1}{c}{6,212}             & \multicolumn{1}{c}{90,573}            & \multicolumn{1}{c}{1}                 & \multicolumn{1}{c}{16,595}            & 442,727           \\ 
\toprule
\multicolumn{7}{c}{\textbf{Word Semantic and Analogy Datasets}}                                                                                                                                                                                                          \\ \toprule
\multicolumn{1}{l}{\textbf{Dataset}}    & \multicolumn{2}{c}{\textbf{No. of unique \textit{V}}}                            & \multicolumn{2}{c}{\textbf{No. of \textit{V}} \textbf{ overlap with CSKG}}                 & \multicolumn{2}{c}{\textbf{Percentage overlap (\%)}}      \\ \hline
\multicolumn{1}{l}{Google Analogy \cite{Mikolov2013EfficientEO}}      & \multicolumn{2}{c}{919}                                                        & \multicolumn{2}{c}{906}                                                        & \multicolumn{2}{c}{98.50}                                 \\ 
\multicolumn{1}{l}{MSR Analogy \cite{mikolov-etal-2013-linguistic}}         & \multicolumn{2}{c}{982}                                                        & \multicolumn{2}{c}{869}                                                        & \multicolumn{2}{c}{88.50}                                 \\ 
\multicolumn{1}{l}{MEN \cite{10.5555/2655713.2655714}}                 & \multicolumn{2}{c}{751}                                                        & \multicolumn{2}{c}{751}                                                        & \multicolumn{2}{c}{100.00}                                   \\ 
\multicolumn{1}{l}{MTruk \cite{10.1145/2339530.2339751}}               & \multicolumn{2}{c}{499}                                                        & \multicolumn{2}{c}{499}                                                        & \multicolumn{2}{c}{100.00}                                   \\ 
\multicolumn{1}{l}{WS353 \cite{finkelstein-2001-placing}}               & \multicolumn{2}{c}{437}                                                        & \multicolumn{2}{c}{437}                                                        & \multicolumn{2}{c}{100.00}                                   \\ 
\multicolumn{1}{l}{RG65 \cite{10.1145/365628.365657}}                & \multicolumn{2}{c}{48}                                                         & \multicolumn{2}{c}{48}                                                         & \multicolumn{2}{c}{100.00}                                   \\ 
\multicolumn{1}{l}{RW \cite{pilehvar-etal-2018-card}}                  & \multicolumn{2}{c}{2951}                                                       & \multicolumn{2}{c}{2926}                                                       & \multicolumn{2}{c}{99.00}                                    \\ 
\multicolumn{1}{l}{SimLex99 \cite{hill-etal-2015-simlex}}            & \multicolumn{2}{c}{1028}                                                       & \multicolumn{2}{c}{1028}                                                       & \multicolumn{2}{c}{100.00}                                   \\ 
\toprule
\multicolumn{1}{l}{Total unique \textit{V} overlap with CSKG across all datasets}          & 
\multicolumn{2}{c}{-}                                                       & \multicolumn{2}{c}{5703}                                                       & \multicolumn{2}{c}{-}                                \\ 
\bottomrule
\end{tabular}
}
\end{center}
\caption{\textbf{Dataset statistics.} \textit{V} is the collection of node set and \textit{E} is the edge set.}
\label{graphstats}
\end{table*}

\section{Graph Embeddings -- Definitions, Preliminaries and Background}
\label{sec:GE}

\textbf{Graph:} A graph $G=(V,E)$ is a collection of node set, $V=\{v_i|i=1,...,n\}$ and edge set, $E \subseteq V \times V$. In this paper, we are assessing a directed and unweighted graph such that the edge weight is uniformly 1. The adjacency matrix $W$ of a graph $G$ is denoted as:

\begin{equation}
    W_i{_j} = \left\{ \begin{array}{cl}
    1 & if (v_i, v_j) \in E \\
    0 & otherwise
    \end{array} \right.
\end{equation}

\noindent \textbf{Graph embedding:} Given a graph $G$, a graph embedding, $Y$ is a mapping $f : v_i \rightarrow y_i \in R^d \, \forall \, i \in [n]$ such that $d \ll |V|$ where $d$ is the dimension size and the scoring function $f$ maps each node to a low-dimensional feature vector space to preserve the topological structure and semantic information of the graph $G$. That is, if there exists a link between $v_i$ and $v_j$, the corresponding embeddings $y_i$ and $y_j$ should be close to each other in the projected vector space.

\subsection{Families of Graph Embedding Algorithms}
\label{sec:GE_algo}
Numerous GE algorithms introduced in the past decade can be categorized into three families \cite{goyal2018graph}: (i) factorization (e.g., Locally Linear Embedding, Laplacian Eigenmaps, HOPE), (ii) random walk (e.g., Node2vec), and (iii) deep learning-based (e.g., SDNE). Given the applicability of GEs in various downstream applications \cite{ameer2019techniques, goyal2018graph}, we are interested in assessing which family of GE algorithms is better at preserving the topological structure versus information between individual nodes. Hence, we select representative algorithms from each family in our evaluation, and the following provides a background of these GE algorithms.

\subsubsection{Family 1: Factorization-based Algorithms.}
Factorization-based algorithms represent the connections between nodes in the form of a matrix and generate embeddings by factorizing the matrix based on different matrix properties. 

\noindent \textbf{\textit{Locally Linear Embedding (LLE)}} \cite{roweis2000nonlinear}: This approach assumes a linear combination of the node's neighbors and is designed to preserve first-order proximity (i.e., one hop). It recovers the embedding $Y^{N \times d}$ from the locally linear fits by minimizing $\phi(Y) = \sum_{i}\big|Y_{i} - \small{\sum_{j}}W_{ij}Y_{j}\big|^{2}$. However, it is not scalable due to its complexity of $O(V^2)$ where $V$ is the number of vertices. 

\noindent \textbf{\textit{Laplacian Eigenmaps (LAP)}} \cite{belkin2001laplacian}: This approach preserves the first-order proximity of a network structure by minimizing $\phi(Y) = \frac{1}{2}\sum_{i,j}||Y_{i} - Y_{j}||^{2}W_{ij}$. Similar to LLE, it has a complexity of $O(V^2)$. 

\noindent \textbf{\textit{High-Order Proximity-preserved Embedding (HOPE)}} \cite{ou2016asymmetric}: This approach preserves asymmetric transitivity and higher order proximity by minimizing $||\textbf{S} - \textbf{Y}_{s}{\textbf{Y}^{T}_{t}}||^2_{F}$, where $S$ is the similarity matrix. Asymmetric transitivity describes the correlation among directed edges in the graph. Asymmetric transitivity between nodes $u$ and $v$ states that if there is a directed path from $u$ to $v$, there is likely a directed edge from $u$ to $v$. HOPE uses generalized Singular Value Decomposition (SVD) \cite{van1976generalizing} to obtain the embedding efficiently with a linear time complexity with respect to the number of edges.
The downside of this GE family is that they are, in general, not capable of approximating an arbitrary function nor structural equivalence (i.e., the extent to which two nodes are connected to the same others) unless explicitly designed into their learning function $\phi(Y)$.

\subsubsection{Family 2: Random Walk-based Algorithms.} 
Random walking is a popular approach to approximate graph properties such as betweenness \cite{newman2005measure} and similarities \cite{fouss2007random} between nodes. 

\noindent \textbf{\textit{Node2Vec}} \cite{grover2016node2vec}: This approach captures network structure by preserving the network neighborhood of a node by maximizing the probability of occurrence of subsequent nodes in fixed-length random walks. Nodes that commonly appear together are embedded closely in the embedding space. It employs a biased-random neighborhood sampling strategy that explores the neighborhoods in a breath-first search (BFS) and depth-first search (DFS) fashion and subsequently feeds them to the Skip-Gram model. Unlike factorization methods, the mixture of community and structural equivalences can be approximated by varying the random walk parameters.

\subsubsection{Family 3: Deep Learning-based Algorithms.}
Deep learning research has seen increasing applications of deep neural networks on graphs due to their ability to approximate a wide range of functions following the universal approximation theorem \cite{hornik1990universal}. One such application is deep auto-encoders \cite{wang2016structural} due to their ability to model non-linearity in graphs. 

\noindent
\textbf{\textit{Structural Deep Network Embeddings (SDNE)}} \cite{wang2016structural}: This approach is a semi-supervised model which uses a coupled deep auto-encoder to embed graphs. It uses highly non-linear functions to capture the non-linearity in network structure by jointly optimizing the first-order and second-order proximities. It consists of an unsupervised part where an autoencoder is employed to embed the nodes such that the reconstruction error is minimized and a supervised part based on Laplacian Eigenmaps \cite{belkin2001laplacian} where a penalty is applied when similar nodes are mapped far from one another in the embedding space. The trained weights of the auto-encoder can be interpreted as the representation of the structure of the graph. 

\cite{goyal2018graph, liu2019much} explores how each GE family represents the global structure at the graph level; however, the degree to which they preserve the structural properties and information at the \textit{node} level based on different levels of order proximity (i.e., number of hops) has not been explored. This is one of the main motivations for our research.

\section{Graph Embedding Assessment Through Reconstruction}
\label{sec:Recon}

We can interpret GE as a representation that encapsulates graphical data: topological structure and local information between nodes. Thus, a ``good" GE is expected to accurately reconstruct the graph (Figure \ref{recon}) and retain the underlying semantics.

\subsection{Topological Structure}
 While there are various methods to evaluate how much graph structure is preserved by embeddings \cite{liu2019much}, we are interested in assessing the notion of nodes that are structurally similar or connected in the graph close together in the vector space. Hence, we opt for the reconstruction assessment based on node proximity \cite{goyal2018graph}. For each GE algorithm, we first reconstruct the adjacency matrix, $W$, based on the proximity of nodes from the generated embeddings. We then rank pairs of nodes according to their normalized proximity (edge weight) score (including only the edge weight greater than the set threshold of \textit{0.5}) and calculate the reconstruction precision based on the ratio of actual links in top $k$ predictions. Figure \ref{arch} and algorithm \ref{alg:Recon} illustrate the process of reconstructing the original graph, $G$ from a trained GE, $Y$, with Node2Vec.

\noindent \textbf{Evaluation Metric}: We use Precision at $k$ (Prec@$k$) and mean Average Precision (mAP)  for evaluating the graph reconstruction. Prec@$k$ is the fraction of correct predictions in top $k$ predictions. It is defined as $Prec@k = \frac{|E_{pred}(1:k) \cap E_{obs}|}{k}$, where $E_{pred}(1:k)$ are the top $k$ predictions and $E_{obs}$ are the observed edges. The $k$ used in our experiments is fractionalized over the total number of nodes, $V$ (i.e., 0.1 denotes 10\% of the reconstructed graph). mAP computes the average precision over all nodes and is defined as $mAP =\frac{\sum_i AP(i)}{|V|}$.

\begin{algorithm}[tb]
\small
\caption{Original Graph Reconstruction from Graph Embedding} 
\label{alg:Recon}
\textbf{Input:} The original graph, $G$ and the trained GE, $Y$ \\
\textbf{Output:} The predicted edge list, $E_{pred}$\\
\textbf{GE Algorithm:} Node2Vec
\begin{algorithmic}[1]
\STATE Reconstruct the adjacency matrix, $W_{recon}$ from $Y$ by computing the node proximity score (edge weight), $W_{recon_{ij}}$ where
\begin{equation}
    W_{recon_{ij}} = norm(y_i \cdot y_j)
\end{equation}
\STATE Get the $E_{pred}$ from $W_{recon}$ where  $E_{pred} = (v_i, v_j, W_{recon_{ij}})$ if $W_{recon_{ij}} \geq$ threshold

\end{algorithmic}
\end{algorithm}

\begin{table*}[ht!]
\begin{center}
\centering
\scalebox{0.65} {
\begin{tabular}{ccccccccc} 
\toprule
\textbf{GE Algorithm}     & \textbf{Hop} & \textbf{mAP}                   & \textbf{Prec@0.1} & \textbf{Prec@0.2} & \textbf{Prec@0.4} & \textbf{Prec@0.6} & \textbf{Prec@0.8} & \textbf{Prec@1.0}  \\ 
\toprule
\multirow{3}{*}{Node2vec} & 1-hop        & 0.53                           & 0.72              & 0.65              & 0.67              & 0.37              & 0.38              & 0.38               \\ 
                          & 2-hop        & 0.17                           & 0.25              & 0.18              & 0.19              & 0.19              & 0.20              & 0.21               \\ 
                          & 3-hop        & 0.16                           & 0.24              & 0.17              & 0.16              & 0.14              & 0.13              & 0.13               \\ 
\midrule
\multirow{3}{*}{HOPE}     & 1-hop        & \textbf{0.92} & 1.00              & 0.99              & 0.95              & 0.91              & 0.85              & 0.83               \\ 
                          & 2-hop        & 0.42                           & 0.89              & 0.43              & 0.38              & 0.30              & 0.27              & 0.24               \\ 
                          & 3-hop        & 0.30                           & 0.99              & 0.24              & 0.18              & 0.14              & 0.11              & 0.10               \\ 
\midrule
\multirow{3}{*}{SDNE}     & 1-hop        & 0.67                           & 0.73              & 0.69              & 0.69              & 0.66              & 0.63              & 0.63               \\ 
                          & 2-hop        & \textbf{0.54} & 0.52              & 0.34              & 0.34              & 0.33              & 0.28              & 0.25               \\ 
                          & 3-hop        & \textbf{0.35} & 0.71              & 0.53              & 0.34              & 0.23              & 0.18              & 0.14               \\ 
\midrule
\multirow{3}{*}{LAP}      & 1-hop        & 0.40                           & 0.55              & 0.27              & 0.38              & 0.42              & 0.41              & 0.36               \\ 
                          & 2-hop        & 0.38                           & 0.69              & 0.31              & 0.34              & 0.33              & 0.33              & 0.30               \\ 
                          & 3-hop        & 0.14                           & 0.51              & 0.22              & 0.18              & 0.13              & 0.11              & 0.09               \\ 
\midrule
\multirow{3}{*}{LLE}      & 1-hop        & 0.71                           & 1.00              & 0.89              & 0.83              & 0.54              & 0.51              & 0.50               \\
                          & 2-hop        & 0.20                           & 0.83              & 0.45              & 0.36              & 0.30              & 0.23              & 0.21               \\ 
                          & 3-hop        & 0.13                           & 0.48              & 0.20              & 0.16              & 0.14              & 0.11              & 0.10               \\
\bottomrule
\end{tabular}
}
\end{center}
\caption{\textbf{mAP and precision metrics of various GE algorithms on the CSKG subgraphs reconstruction.} The higher, the better. Prec@\{0.1, 0.2, 0.4, 0.6, 0.8, 1.0\} denotes precision at \{10\%, 20\%, 40\%, 60\%, 80\%, 100\%\} of the reconstructed subgraph.}
\label{MetricValues}
\end{table*}

\begin{figure*}[t]
\centering
\includegraphics[width=0.75\textwidth]{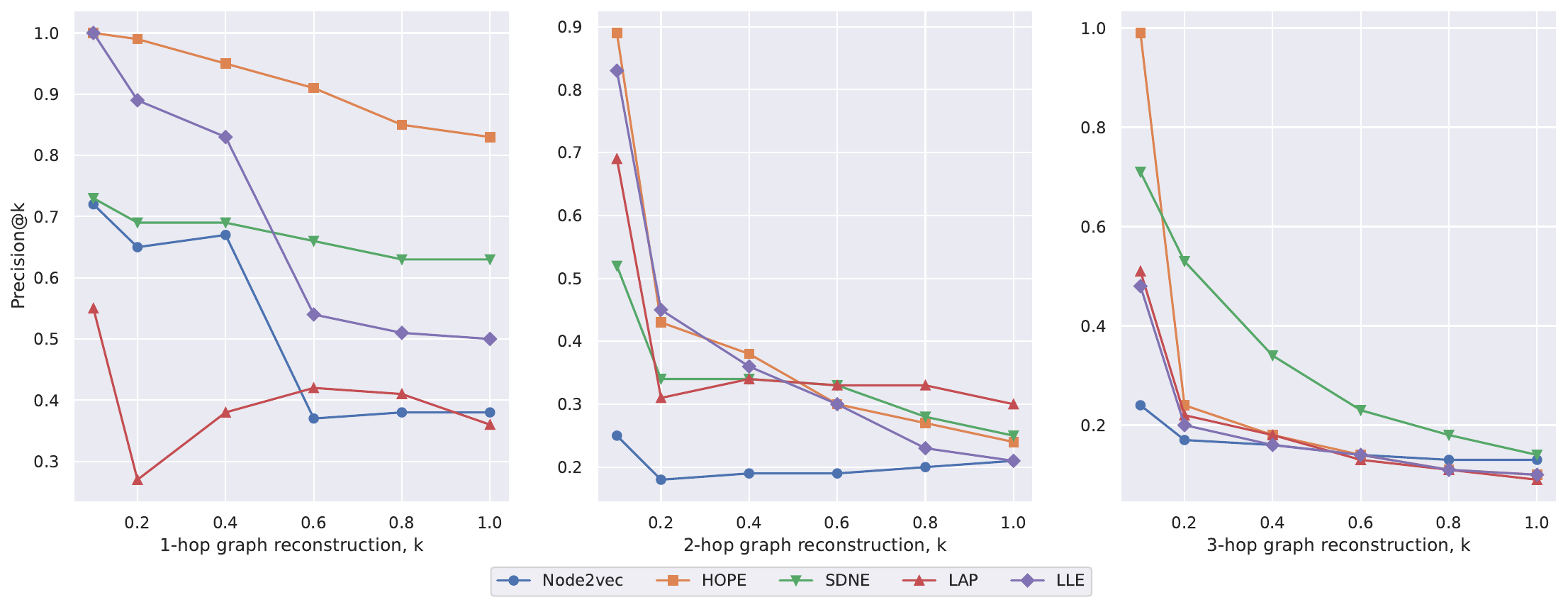} 

\caption{\textbf{Precision@\textit{k} for 1, 2, and 3-hop graph reconstructions.} Prec@\textit{k} is the fraction of correct predictions in top \textit{k} predictions where \textit{k} is fractionalized over the total number of nodes, \textit{V} (i.e., 0.1 denotes 10\% of the reconstructed subgraph).}
\label{res}
\end{figure*}

\subsection{Semantic Information}
To assess the learned semantic information between nodes, we adopt the word embeddings evaluation framework proposed by \cite{jastrzebski2017evaluate}, which consists of an array of datasets divided into two evaluation categories: \textit{Similarity} and \textit{Analogy}. The \textit{Similarity} datasets comprise pairs of words and assigned a mean rank by human annotators, while the \textit{Analogy} datasets consist of quadruples: two pairs of words bounded by an intrinsic relation (e.g., play-played, make-made).

\noindent
\textbf{\textit{Similarity:}} Multimodal Distributional Semantics (MEN) \cite{10.5555/2655713.2655714}, MTruk \cite{10.1145/2339530.2339751}, WS353 (Word Similarity) \cite{finkelstein-2001-placing}, RG65 \cite{10.1145/365628.365657}, Rare Words (RW) \cite{pilehvar-etal-2018-card}, and Similarity Estimation (SimLex99) \cite{hill-etal-2015-simlex}

\noindent
\textbf{\textit{Analogy:}} Google Analogy \cite{Mikolov2013EfficientEO} and MSR Analogy \cite{mikolov-etal-2013-linguistic}

\noindent
\textbf{Evaluation Metric}: We assume the notion that semantically similar nodes are positioned closely to one another in the vector space. We use Euclidean Distance to compute the distance, $D$, between two node embeddings, $y_1$ and $y_2$, which is expressed as 
$D\left(y_1,y_2\right) = \sqrt {\sum _{i=1}^{n}  \left( y_{2_i}-y_{1_i}\right)^2 }$ where $n$ is the size of the vector.

%% file: 3_results.tex

\begin{table*}[h]
    \begin{center}
    \scalebox{0.65}{
        \begin{tabular}{cccccccc}
        \toprule
        \multirow{2}{*}{\textbf{GE Algorithm}} & \multirow{2}{*}{\textbf{Hop}} & \multicolumn{3}{c}{\textbf{Nodes, V}}                                                                                                                                                                                                                                     & \multicolumn{3}{c}{\textbf{Edges, E}}                                                                                                                                                                                                                                     \\ \cline{3-8} 
                                       &                               & \multicolumn{1}{c}{\textbf{\begin{tabular}[c]{@{}c@{}}Avg. no. of \\ V\end{tabular}}} & \multicolumn{1}{c}{\textbf{\begin{tabular}[c]{@{}c@{}}Avg. no. of \\ added V\end{tabular}}} & \textbf{\begin{tabular}[c]{@{}c@{}}Avg. no. of \\ missing V\end{tabular}} & \multicolumn{1}{c}{\textbf{\begin{tabular}[c]{@{}c@{}}Avg. no. of \\ E\end{tabular}}} & \multicolumn{1}{c}{\textbf{\begin{tabular}[c]{@{}c@{}}Avg. no. of \\ added E\end{tabular}}} & \textbf{\begin{tabular}[c]{@{}c@{}}Avg. no. of \\ missing E\end{tabular}} \\ \toprule
        \multirow{3}{*}{Node2vec}              & 1-hop                         & \multicolumn{1}{c}{4}                        & \multicolumn{1}{c}{0}                              & 0                                                     & \multicolumn{1}{c}{3}                        & \multicolumn{1}{c}{40}                             & 0                                                     \\ 
                                               & 2-hop                         & \multicolumn{1}{c}{215}                      & \multicolumn{1}{c}{0}                              & 0                                                     & \multicolumn{1}{c}{250}                      & \multicolumn{1}{c}{102,370}                        & 15                                                    \\ 
                                               & 3-hop                         & \multicolumn{1}{c}{6,212}                     & \multicolumn{1}{c}{0}                              & 0                                                     & \multicolumn{1}{c}{16,595}                    & \multicolumn{1}{c}{145,416}                        & 2,265                                                 \\ \midrule
        \multirow{3}{*}{HOPE}                  & 1-hop                         & \multicolumn{1}{c}{4}                        & \multicolumn{1}{c}{0}                              & 0                                                     & \multicolumn{1}{c}{3}                        & \multicolumn{1}{c}{1}                              & 0                                                     \\ 
                                               & 2-hop                         & \multicolumn{1}{c}{215}                      & \multicolumn{1}{c}{0}                              & 0                                                     & \multicolumn{1}{c}{250}                      & \multicolumn{1}{c}{59,889}                         & 15                                                    \\ 
                                               & 3-hop                         & \multicolumn{1}{c}{6,212}                     & \multicolumn{1}{c}{0}                              & 3,352                                                  & \multicolumn{1}{c}{16,595}                    & \multicolumn{1}{c}{110,450}                        & 1                                                     \\ \midrule
        \multirow{3}{*}{SDNE}                  & 1-hop                         & \multicolumn{1}{c}{4}                        & \multicolumn{1}{c}{0}                              & 0                                                     & \multicolumn{1}{c}{3}                        & \multicolumn{1}{c}{10}                             & 1                                                     \\ 
                                               & 2-hop                         & \multicolumn{1}{c}{215}                      & \multicolumn{1}{c}{0}                              & 50                                                    & \multicolumn{1}{c}{250}                      & \multicolumn{1}{c}{50,239}                         & 134                                                   \\ 
                                               & 3-hop                         & \multicolumn{1}{c}{6,212}                     & \multicolumn{1}{c}{0}                              & 1,812                                                  & \multicolumn{1}{c}{16,595}                    & \multicolumn{1}{c}{90,141}                         & 19,905                                                \\ \midrule
        \multirow{3}{*}{LAP}                   & 1-hop                         & \multicolumn{1}{c}{4}                        & \multicolumn{1}{c}{0}                              & 1                                                     & \multicolumn{1}{c}{3}                        & \multicolumn{1}{c}{44}                             & 0                                                     \\ 
                                               & 2-hop                         & \multicolumn{1}{c}{215}                      & \multicolumn{1}{c}{0}                              & 0                                                     & \multicolumn{1}{c}{250}                      & \multicolumn{1}{c}{57,936}                         & 9                                                     \\ 
                                               & 3-hop                         & \multicolumn{1}{c}{6,212}                     & \multicolumn{1}{c}{0}                              & 8                                                     & \multicolumn{1}{c}{16,595}                    & \multicolumn{1}{c}{155,256}                        & 21                                                    \\ \midrule
        \multirow{3}{*}{LLE}                   & 1-hop                         & \multicolumn{1}{c}{4}                        & \multicolumn{1}{c}{0}                              & 0                                                     & \multicolumn{1}{c}{3}                        & \multicolumn{1}{c}{18}                             & 0                                                     \\ 
                                               & 2-hop                         & \multicolumn{1}{c}{215}                      & \multicolumn{1}{c}{0}                              & 0                                                     & \multicolumn{1}{c}{250}                      & \multicolumn{1}{c}{91,056}                         & 13                                                    \\ 
                                               & \multicolumn{1}{l}{3-hop}    & \multicolumn{1}{c}{6,212}                     & \multicolumn{1}{c}{0}                              & 4                                                     & \multicolumn{1}{c}{16,595}                    & \multicolumn{1}{c}{162,860}                        & 7                                                     \\ \bottomrule
        \end{tabular}
    }
    \end{center}
    \caption{\textbf{The average number of added, missing nodes and edges by various GE algorithms during reconstruction.} }
    \label{tab:graph-recon-stats}
\end{table*}

\begin{table*}[ht!]
\begin{center}
\centering
\scalebox{0.62} {
\begin{tabular}{ccccccccccccccccc} 
\toprule
\multirow{2}{*}{\textbf{Dataset}} & \multirow{2}{*}{\textbf{GloVe}} & \multicolumn{3}{c}{\textbf{Node2Vec}}            & \multicolumn{3}{c}{\textbf{HOPE}}                                                                & \multicolumn{3}{c}{\textbf{SDNE}}                & \multicolumn{3}{c}{\textbf{LAP}}                 & \multicolumn{3}{c}{\textbf{LLE}}                  \\ 
\cline{3-17}
                                  &                                 & \textbf{1-hop} & \textbf{2-hop} & \textbf{3-hop} & \textbf{1-hop}                 & \textbf{2-hop}                 & \textbf{3-hop}                 & \textbf{1-hop} & \textbf{2-hop} & \textbf{3-hop} & \textbf{1-hop} & \textbf{2-hop} & \textbf{3-hop} & \textbf{1-hop} & \textbf{2-hop} & \textbf{3-hop}  \\ 
\cmidrule[\heavyrulewidth]{1-17}
Google Analogy                    & 2.09                            & 1.28           & 1.89           & 4.26           & 0.12                           & 0.17                           & 0.08                           & 1.92           & 2.90           & 3.01           & 0.63           & 1.15           & 1.26           & 0.48           & 0.76           & 0.96            \\ 
MSR                               & 0.63                            & 1.26           & 1.79           & 4.20           & 0.10                           & 0.15                           & 0.07                           & 4.30           & 4.69           & 4.96           & 0.72           & 1.21           & 1.33           & 0.18           & 1.06           & 1.25            \\ 
MEN                               & 0.51                            & 1.28           & 1.79           & 4.16           & 0.12                           & 0.17                           & 0.09                           & 4.41           & 5.18           & 5.79           & 0.73           & 1.22           & 1.35           & 0.62           & 0.65           & 0.76            \\ 
MTurk                             & 1.99                            & 1.31           & 1.83           & 4.23           & 0.17                           & 0.18                           & 0.12                           & 1.15           & 1.27           & 1.33           & 0.64           & 1.22           & 1.33           & 0.58           & 1.05           & 1.36            \\ 
WS353                             & 2.65                            & 1.35           & 1.78           & 3.98           & 0.16                           & 0.17                           & 0.16                           & 1.08           & 1.12           & 1.74           & 0.79           & 1.15           & 1.25           & 0.66           & 1.05           & 1.47            \\ 
RG65                              & 0.75                            & 1.28           & 1.79           & 4.41           & 0.15                           & 0.17                           & 0.10                           & 0.63           & 1.73           & 2.05           & 0.74           & 1.18           & 1.29           & 0.58           & 1.07           & 1.28            \\ 
RW                                & 0.96                            & 1.27           & 1.63           & 3.82           & 0.15                           & 0.16                           & 0.14                           & 1.88           & 2.58           & 2.67           & 0.72           & 1.17           & 1.51           & 0.50           & 1.07           & 0.94            \\ 
SIMLEX99                          & 2.31                            & 1.29           & 1.75           & 3.89           & 0.16                           & 0.17                           & 0.12                           & 2.76           & 4.82           & 4.93           & 0.67           & 1.20           & 0.92           & 0.63           & 1.00           & 1.14            \\ 
\cmidrule[\heavyrulewidth]{1-17}
Average                           & 1.49                            & 1.29           & 1.78           & 4.12           & \textbf{0.14} & \textbf{0.17} & \textbf{0.11} & 2.26           & 3.04           & 3.31           & 0.70           & 1.19           & 1.28           & 0.53           & 0.96           & 1.14            \\
\bottomrule
\end{tabular}
}
\end{center}
\caption{\textbf{Word semantic and analogy tests based on pairwise Euclidean distance measure.} The lower, the better.}
\label{EuclideanWordAnalogy}
\end{table*}

%% file: 4_evaluation.tex
\section{Experiments and Analysis}

We re-implement and train the different GE algorithms with GEM \cite{goyal2018graph} on the Commonsense Knowledge Graph (CSKG) \cite{ilievski2021cskg}. The rationale for using CSKG is two-fold: (1) with the increasing research on incorporating relevant knowledge (i.e., in the form of infusing GE trained on extracted subgraphs onto neural networks) on various downstream tasks \cite{kursuncu2019knowledge}, we are interested in assessing the quality of the GEs generated from a popular knowledge graph with varying type of graph properties; and (2) the CSKG corpus contains up to 88.5\% of the vocabularies present in our word semantic and analogy tests (Table \ref{graphstats}), which allows for a wide assessment of the degree of semantic information preserved by the GEs. All experiments are performed on a CentOS Linux 7 system with a hyperthreaded Intel Xeon Platinum 8260 processor with 24 cores and a clock speed of 2.4 GHz, 200 GB RAM, and a Tesla V100 32 GB GPU.

\subsection{Dataset}

CSKG combines seven popular sources into a consolidated representation: ATOMIC, ConceptNet, FrameNet, Roget, Visual Genome, Wikidata, and WordNet \cite{ilievski2021cskg}. It covers a rich spectrum of knowledge ranging from every day to event-centric knowledge and taxonomies to visual knowledge. It is modeled as a \textit{directed}, \textit{unweighted}, and \textit{hyper-relational} graph with 2,160,968 nodes and 6,001,531 edges. In our assessment, we are \textit{not} training and evaluating the GE algorithms on the entirety of the CSKG graph due to (1) limitations in algorithmic scalability and (2) in most downstream applications, extracting only subgraphs relevant to the entities or communities of interest (domain-specific) is generally the strategy for learning semantically relevant embeddings. Instead, for each vocabulary that is present in both CSKG and the word semantic and analogy tests (a total of 5703 total unique vocabularies overlap with CSKG across all datasets), we first retrieve and generate the corresponding subgraphs of size 1, 2, and 3-hop. We then train and evaluate embeddings for these subgraphs with each GE algorithm. Table \ref{graphstats} shows a summary of the dataset statistics.

\subsection{Hyperparameters Selection}
Given the cost, time, and increasing complexity of the GE algorithms in generating 1, 2, and 3-hop subgraphs and their corresponding GEs for each vocabulary (node of interest), we are interested in assessing the different off-the-shelf characteristics and performance of the various GE families with the commonly-used list of hyper-parameters.

\noindent \textbf{Node2Vec:} We use a \textit{context size} of 10; \textit{walk length} of 80; and both \textit{inout} and \textit{return} parameter of 1.

\noindent \textbf{HOPE:} We use the Katz index, which describes the similarity between of $v_i$ and $v_j$ to compute the similarity matrix, $S$ with the attenuation factor, $\beta$ set to 0.01. 

\noindent \textbf{SDNE:} We use 2 hidden layers with 50 and 15 hidden units for the encoder/decoder layer respectively; \textit{alpha} of $1e^{-5}$; \textit{beta} of 5; both L1 and L2 regularization of $1e^{-6}$; \textit{rho} of 0.3; \textit{xeta} of 0.01; and a batch size of 100.

There are no hyperparameters for \textbf{\textit{LAP}} and \textbf{\textit{LLE}}. In addition, the following parameters are set \textit{constant} across all settings whenever applicable: the embedding sizes, \textit{d}, for 1-hop, 2-hop, and 3-hop graphs are set to 2, 64, and 128 respectively with 50 training iterations.

%% file: 5_results_dis.tex
\section{Results}
We report our assessment of the quality of embeddings generated by different families of GE algorithms through graph reconstruction with increasing levels of order proximity in Table \ref{MetricValues} and Figure \ref{res}. Table \ref{tab:graph-recon-stats} shows the average number of added, missing nodes and edges by various GE algorithms during the subgraphs reconstruction with increasing number of hops. Table \ref{EuclideanWordAnalogy} reports the average Euclidean distance measure of the various GEs on the array of word semantic and analogy tests. Next, we summarize our observations along each dimension across all GE algorithms.

\noindent \textbf{1. Understand the effectiveness of graph embeddings in preserving the type of graph properties: (a) topological structure and (b) semantic information between nodes with increasing graph size.} For (a), we observed in Table \ref{MetricValues} that HOPE outperforms others in 1-hop graph reconstruction with the highest mAP of 0.92 with a gradual decrease in performance as the graph is being reconstructed. SDNE demonstrates overall better performance at 2 and 3-hop graph reconstructions with mAP of 0.54 and 0.35 respectively. For (b), Table \ref{EuclideanWordAnalogy} shows that HOPE performs the best across all the word similarity and analogy tests and its performance does not fluctuate with the increasing number of hops (1-3), with the lowest average Euclidean distance of 0.14, 0.17, and 0.11 respectively. On the contrary, SDNE performs the worst with an average distance of 2.26, 3.04, and 3.31 with increasing order of proximity.

\noindent \textbf{2. Which \textit{family} of GE algorithm(s) are \textit{better} at preserving 1(a) versus 1(b).} For 1(a), all families of GE algorithms show a similar \textit{downward} trend in their reconstruction performances with increasing number of hops and graph size (Figure \ref{res}). Nonetheless, deep-learning algorithm (SDNE) demonstrates consistent performance in preserving the global structure of the graph across 1, 2, and 3-hop when compared to the other GE algorithms. On the contrary, from Table \ref{EuclideanWordAnalogy}, we observe that the factorization-based family of GE algorithms (HOPE, LAP, and LLE) is more suited for capturing 1(b) as demonstrated by their consistently low Euclidean distances when compared to random walk (Node2Vec) and deep learning-based (SDNE) as well as GloVe embedding \cite{pennington} across all word similarity and analogy tests.
  
\noindent \textbf{3. The degree of information that is preserved (retained, added, and missed) by the various GE algorithms with increasing \textit{number of hops}}. From Table \ref{tab:graph-recon-stats}, we observe that all GE algorithms suffer from incorrectly added and missing edges. The average number of incorrectly added edges scales exponentially with the increasing number of hops across all GE algorithms (notably with 3-hop). While SDNE (which shows the best mAP score for preserving the topological structure for 2 and 3-hop subgraphs) has the lowest average number of added edges, it also has the highest number of missing edges. HOPE follows second, and Node2vec is the worst-performing GE algorithm overall when compared to LAP and LLE, which also perform relatively poorly in 3-hop graph reconstruction.

%% file: 6_discussion_futurework.tex
\section{Discussion and Future Work}

\textbf{Findings.} We learn that different families of GE algorithms capture different information and there is no "one size fits all" approach to preserving the full length of the original graph properties. The best performance (mAP) for 1-hop, 2-hop, and 3-hop reconstructions is only capturing 0.92 (HOPE), 0.54 (SDNE), and 0.35 (SDNE) of the original graph information. Dissecting the information that was supposedly \textit{learned}, we observe that these GEs are creating additional links between nodes and missing edges, which can plausibly be translated to factually incorrect knowledge, thereby limiting the benefits of GEs.

\noindent \textbf{Significance.} Suppose we assume the performance improvement based on extrinsic evaluations by current deep neural networks on downstream applications is attributed to the use of GEs (Makarov et al. 2021), it is critical that the GEs are preserving the right graph structure and semantics. In this work, we shed light on the effectiveness and shortcomings of GEs. We hope these insights encourage new research avenues on approaches to better improve the graph representation and learning, to which, an auto-encoder/decoder framework of generating meta-embeddings \cite{bollegala2018learning} based on a combination of multiple source embeddings to produce more accurate and complete GEs warrants a promising direction.

\noindent \textbf{Limitation and Future Work.} As mentioned earlier, this work omits the reconstruction evaluation of typed/ labeled relations. We formulate the task as a graph reconstruction task instead of a link prediction task. Our focus is rather on the reconstruction of existing links between nodes rather than link discovery, to which we based our findings that the addition of new and/or missing links contribute to the inaccurate representation of the original graph. We acknowledge the other form of GE, the Knowledge Graph Embeddings (KGEs) which treats typed relationships as first-class citizens. Performing a comparative analysis between network embeddings (used in this work) and KGEs was beyond the scope of this investigation but is part of our future work. In addition, we plan to improve the current performance of GEs, e.g., by developing novel meta-embedding techniques that consider both structure and semantic information.

%% file: 7_conclusion.tex
\section{Conclusion}

In summary, we proposed RESTORE, which is an intrinsic evaluation framework that aims to assess the quality and effectiveness of GEs in retaining the original graph topological structure and semantic information through reconstruction. Understanding these will help identify the deficiency and yield insights into these GEs when vectorizing graphs in terms of preserving the relevant knowledge or learning incorrect knowledge (i.e., incorrectly added and missing nodes as well as edges). Particularly, we show that deep learning-based GEs are better at preserving the global topological structure and factorization-based GEs are more suited for capturing the semantic information. Nonetheless, the modest performance of these GEs leaves room for further research avenues on better graph representation learning.

%% file: aaai24.bbl
\begin{thebibliography}{34}
\providecommand{\natexlab}[1]{#1}

\bibitem[{Ahmed et~al.(2013)Ahmed, Shervashidze, Narayanamurthy, Josifovski,
  and Smola}]{ahmed2013distributed}
Ahmed, A.; Shervashidze, N.; Narayanamurthy, S.; Josifovski, V.; and Smola,
  A.~J. 2013.
\newblock Distributed large-scale natural graph factorization.
\newblock In \emph{Proceedings of the 22nd international conference on World
  Wide Web}, 37--48.

\bibitem[{Ameer et~al.(2019)Ameer, Hanif, Talib, Sarwar, Khan, Zulfiqar, and
  Riasat}]{ameer2019techniques}
Ameer, F.; Hanif, M.~K.; Talib, R.; Sarwar, M.~U.; Khan, Z.; Zulfiqar, K.; and
  Riasat, A. 2019.
\newblock Techniques, Tools and Applications of Graph Analytic.
\newblock \emph{International Journal of Advanced Computer Science and
  Applications}, 10(4).

\bibitem[{Belkin and Niyogi(2001)}]{belkin2001laplacian}
Belkin, M.; and Niyogi, P. 2001.
\newblock Laplacian eigenmaps and spectral techniques for embedding and
  clustering.
\newblock \emph{Advances in neural information processing systems}, 14.

\bibitem[{Bollegala and Bao(2018)}]{bollegala2018learning}
Bollegala, D.; and Bao, C. 2018.
\newblock Learning word meta-embeddings by autoencoding.
\newblock In \emph{Proceedings of the 27th international conference on
  computational linguistics}, 1650--1661.

\bibitem[{Bruni, Tran, and Baroni(2014)}]{10.5555/2655713.2655714}
Bruni, E.; Tran, N.~K.; and Baroni, M. 2014.
\newblock Multimodal Distributional Semantics.
\newblock \emph{J. Artif. Int. Res.}, 49(1): 1–47.

\bibitem[{Finkelstein et~al.(2001)Finkelstein, Gabrilovich, Matias, Rivlin,
  Solan, Wolfman, and Ruppin}]{finkelstein-2001-placing}
Finkelstein, L.; Gabrilovich, E.; Matias, Y.; Rivlin, E.; Solan, Z.; Wolfman,
  G.; and Ruppin, E. 2001.
\newblock Placing search in context: the concept revisited.
\newblock In Shen, V.~Y.; Saito, N.; Lyu, M.~R.; and Zurko, M.~E., eds.,
  \emph{Proceedings of the Tenth International World Wide Web Conference, {WWW}
  10, Hong Kong, China, May 1-5, 2001}, 406--414. {ACM}.

\bibitem[{Fouss et~al.(2007)Fouss, Pirotte, Renders, and
  Saerens}]{fouss2007random}
Fouss, F.; Pirotte, A.; Renders, J.-M.; and Saerens, M. 2007.
\newblock Random-walk computation of similarities between nodes of a graph with
  application to collaborative recommendation.
\newblock \emph{IEEE Transactions on knowledge and data engineering}, 19(3):
  355--369.

\bibitem[{Fu et~al.(2021)Fu, Mao, Wang, Lin, Zhang, Zhan, Sun, and
  Li}]{fu2021ts}
Fu, K.; Mao, T.; Wang, Y.; Lin, D.; Zhang, Y.; Zhan, J.; Sun, X.; and Li, F.
  2021.
\newblock TS-Extractor: large graph exploration via subgraph extraction based
  on topological and semantic information.
\newblock \emph{Journal of visualization}, 24(1): 173--190.

\bibitem[{Goyal and Ferrara(2018)}]{goyal2018graph}
Goyal, P.; and Ferrara, E. 2018.
\newblock Graph embedding techniques, applications, and performance: A survey.
\newblock \emph{Knowledge-Based Systems}, 151: 78--94.

\bibitem[{Grover and Leskovec(2016)}]{grover2016node2vec}
Grover, A.; and Leskovec, J. 2016.
\newblock node2vec: Scalable feature learning for networks.
\newblock In \emph{Proceedings of the 22nd ACM SIGKDD international conference
  on Knowledge discovery and data mining}, 855--864.

\bibitem[{Halawi et~al.(2012)Halawi, Dror, Gabrilovich, and
  Koren}]{10.1145/2339530.2339751}
Halawi, G.; Dror, G.; Gabrilovich, E.; and Koren, Y. 2012.
\newblock Large-Scale Learning of Word Relatedness with Constraints.
\newblock In \emph{Proceedings of the 18th ACM SIGKDD International Conference
  on Knowledge Discovery and Data Mining}, KDD '12, 1406–1414. New York, NY,
  USA: Association for Computing Machinery.
\newblock ISBN 9781450314626.

\bibitem[{Hill, Reichart, and Korhonen(2015)}]{hill-etal-2015-simlex}
Hill, F.; Reichart, R.; and Korhonen, A. 2015.
\newblock {S}im{L}ex-999: Evaluating Semantic Models With (Genuine) Similarity
  Estimation.
\newblock \emph{Computational Linguistics}, 41(4): 665--695.

\bibitem[{Hornik, Stinchcombe, and White(1990)}]{hornik1990universal}
Hornik, K.; Stinchcombe, M.; and White, H. 1990.
\newblock Universal approximation of an unknown mapping and its derivatives
  using multilayer feedforward networks.
\newblock \emph{Neural networks}, 3(5): 551--560.

\bibitem[{Ilievski, Szekely, and Zhang(2021)}]{ilievski2021cskg}
Ilievski, F.; Szekely, P.; and Zhang, B. 2021.
\newblock Cskg: The commonsense knowledge graph.
\newblock In \emph{European Semantic Web Conference}, 680--696. Springer.

\bibitem[{Jastrzebski, Le{\'s}niak, and
  Czarnecki(2017)}]{jastrzebski2017evaluate}
Jastrzebski, S.; Le{\'s}niak, D.; and Czarnecki, W.~M. 2017.
\newblock How to evaluate word embeddings? on importance of data efficiency and
  simple supervised tasks.
\newblock \emph{arXiv preprint arXiv:1702.02170}.

\bibitem[{Kipf and Welling(2016)}]{kipf2016semi}
Kipf, T.~N.; and Welling, M. 2016.
\newblock Semi-supervised classification with graph convolutional networks.
\newblock \emph{arXiv preprint arXiv:1609.02907}.

\bibitem[{Kursuncu, Gaur, and Sheth(2019)}]{kursuncu2019knowledge}
Kursuncu, U.; Gaur, M.; and Sheth, A. 2019.
\newblock Knowledge infused learning (k-il): Towards deep incorporation of
  knowledge in deep learning.
\newblock \emph{arXiv preprint arXiv:1912.00512}.

\bibitem[{Liu et~al.(2019)Liu, Zhuang, Murata, Kim, and
  Kertkeidkachorn}]{liu2019much}
Liu, X.; Zhuang, C.; Murata, T.; Kim, K.-S.; and Kertkeidkachorn, N. 2019.
\newblock How much topological structure is preserved by graph embeddings?
\newblock \emph{Computer Science and Information Systems}, 16(2): 597--614.

\bibitem[{Makarov et~al.(2021)Makarov, Kiselev, Nikitinsky, and
  Subelj}]{makarov2021survey}
Makarov, I.; Kiselev, D.; Nikitinsky, N.; and Subelj, L. 2021.
\newblock Survey on graph embeddings and their applications to machine learning
  problems on graphs.
\newblock \emph{PeerJ Computer Science}, 7: e357.

\bibitem[{Mikolov et~al.(2013{\natexlab{a}})Mikolov, Chen, Corrado, and
  Dean}]{mikolov2013efficient}
Mikolov, T.; Chen, K.; Corrado, G.; and Dean, J. 2013{\natexlab{a}}.
\newblock Efficient estimation of word representations in vector space.
\newblock \emph{arXiv preprint arXiv:1301.3781}.

\bibitem[{Mikolov et~al.(2013{\natexlab{b}})Mikolov, Chen, Corrado, and
  Dean}]{Mikolov2013EfficientEO}
Mikolov, T.; Chen, K.; Corrado, G.~S.; and Dean, J. 2013{\natexlab{b}}.
\newblock Efficient Estimation of Word Representations in Vector Space.
\newblock In \emph{ICLR}.

\bibitem[{Mikolov, Yih, and Zweig(2013)}]{mikolov-etal-2013-linguistic}
Mikolov, T.; Yih, W.-t.; and Zweig, G. 2013.
\newblock Linguistic Regularities in Continuous Space Word Representations.
\newblock In \emph{Proceedings of the 2013 Conference of the North {A}merican
  Chapter of the Association for Computational Linguistics: Human Language
  Technologies}, 746--751. Atlanta, Georgia: Association for Computational
  Linguistics.

\bibitem[{Newman(2005)}]{newman2005measure}
Newman, M.~E. 2005.
\newblock A measure of betweenness centrality based on random walks.
\newblock \emph{Social networks}, 27(1): 39--54.

\bibitem[{Ou et~al.(2016)Ou, Cui, Pei, Zhang, and Zhu}]{ou2016asymmetric}
Ou, M.; Cui, P.; Pei, J.; Zhang, Z.; and Zhu, W. 2016.
\newblock Asymmetric transitivity preserving graph embedding.
\newblock In \emph{Proceedings of the 22nd ACM SIGKDD international conference
  on Knowledge discovery and data mining}, 1105--1114.

\bibitem[{Pennington, Socher, and Manning(2014)}]{pennington}
Pennington, J.; Socher, R.; and Manning, C. 2014.
\newblock {G}lo{V}e: Global Vectors for Word Representation.
\newblock In \emph{Proceedings of the 2014 Conference on Empirical Methods in
  Natural Language Processing ({EMNLP})}, 1532--1543. Doha, Qatar: Association
  for Computational Linguistics.

\bibitem[{Perozzi, Al-Rfou, and Skiena(2014)}]{perozzi2014deepwalk}
Perozzi, B.; Al-Rfou, R.; and Skiena, S. 2014.
\newblock Deepwalk: Online learning of social representations.
\newblock In \emph{Proceedings of the 20th ACM SIGKDD international conference
  on Knowledge discovery and data mining}, 701--710.

\bibitem[{Pilehvar et~al.(2018)Pilehvar, Kartsaklis, Prokhorov, and
  Collier}]{pilehvar-etal-2018-card}
Pilehvar, M.~T.; Kartsaklis, D.; Prokhorov, V.; and Collier, N. 2018.
\newblock Card-660: {C}ambridge Rare Word Dataset - a Reliable Benchmark for
  Infrequent Word Representation Models.
\newblock In \emph{Proceedings of the 2018 Conference on Empirical Methods in
  Natural Language Processing}, 1391--1401. Brussels, Belgium: Association for
  Computational Linguistics.

\bibitem[{Ribeiro et~al.(2021)Ribeiro, Paredes, Silva, Aparicio, and
  Silva}]{ribeiro2021survey}
Ribeiro, P.; Paredes, P.; Silva, M.~E.; Aparicio, D.; and Silva, F. 2021.
\newblock A survey on subgraph counting: concepts, algorithms, and applications
  to network motifs and graphlets.
\newblock \emph{ACM Computing Surveys (CSUR)}, 54(2): 1--36.

\bibitem[{Roweis and Saul(2000)}]{roweis2000nonlinear}
Roweis, S.~T.; and Saul, L.~K. 2000.
\newblock Nonlinear dimensionality reduction by locally linear embedding.
\newblock \emph{science}, 290(5500): 2323--2326.

\bibitem[{Rubenstein and Goodenough(1965)}]{10.1145/365628.365657}
Rubenstein, H.; and Goodenough, J.~B. 1965.
\newblock Contextual Correlates of Synonymy.
\newblock \emph{Commun. ACM}, 8(10): 627–633.

\bibitem[{Tang et~al.(2015)Tang, Qu, Wang, Zhang, Yan, and Mei}]{tang2015line}
Tang, J.; Qu, M.; Wang, M.; Zhang, M.; Yan, J.; and Mei, Q. 2015.
\newblock Line: Large-scale information network embedding.
\newblock In \emph{Proceedings of the 24th international conference on world
  wide web}, 1067--1077.

\bibitem[{Van~Loan(1976)}]{van1976generalizing}
Van~Loan, C.~F. 1976.
\newblock Generalizing the singular value decomposition.
\newblock \emph{SIAM Journal on numerical Analysis}, 13(1): 76--83.

\bibitem[{Wang, Cui, and Zhu(2016)}]{wang2016structural}
Wang, D.; Cui, P.; and Zhu, W. 2016.
\newblock Structural deep network embedding.
\newblock In \emph{Proceedings of the 22nd ACM SIGKDD international conference
  on Knowledge discovery and data mining}, 1225--1234.

\bibitem[{Xu, Torres, and Eliassi-Rad(2017)}]{xucloser}
Xu, J.; Torres, L.; and Eliassi-Rad, T. 2017.
\newblock A Closer Look at Graph Embedding for Graph Reconstruction.

\end{thebibliography}
